\newif\ifanon
\anonfalse

\ifanon
  \documentclass[letterpaper, 10 pt, conference]{ieeeconf}
\else
  \documentclass[letterpaper, 10pt, journal, twoside]{IEEEtran}
\fi

\ifanon
  \IEEEoverridecommandlockouts   
\fi

\usepackage{comment}
\ifanon
  \includecomment{anonblock}\excludecomment{realblock}
\else
  \excludecomment{anonblock}\includecomment{realblock}
\fi

\ifanon
  \newcommand{\parstart}[2]{#1#2}
\else
  \newcommand{\parstart}[2]{\IEEEPARstart{#1}{#2}}
\fi

\ifanon\else
  \usepackage{eso-pic}
  \AddToShipoutPictureBG{%
    \ifnum\value{page}=1
      \put(\LenToUnit{.5\paperwidth},\LenToUnit{0.35in}){%
        \makebox(0,0){\fbox{\parbox{0.92\textwidth}{\centering\footnotesize
          This work has been submitted to the IEEE for possible publication. \\
          Copyright may be transferred without notice, after which this
          version may no longer be accessible.}}}%
      }%
    \fi
  }
\fi

\usepackage{amsmath,amsfonts}
\usepackage{algorithmic}
\usepackage{algorithm}
\usepackage{array}
\usepackage[caption=false,font=normalsize,labelfont=sf,textfont=sf]{subfig}
\usepackage{textcomp}
\usepackage{stfloats}
\usepackage{url}
\usepackage{verbatim}
\usepackage{graphicx}
\usepackage{cite}
\usepackage[hidelinks,hypertexnames=false]{hyperref}
\usepackage{amsbsy}
\usepackage{romannum}
\usepackage{booktabs}
\usepackage{threeparttable}
\usepackage{multirow}
\usepackage{colortbl}
\usepackage{bm}
\usepackage[table, svgnames]{xcolor}
\usepackage{pifont}
\usepackage{todonotes}
\presetkeys{todonotes}{color=blue!20}{}

\definecolor{tablegrey}{HTML}{EFEFEF}
\definecolor{crimson}{HTML}{DC143C}
\definecolor{mydarkgrey}{HTML}{555555}
\definecolor{oracleblue}{RGB}{232,241,250}
\definecolor{oraclecyan}{RGB}{230,245,247}
\definecolor{BaselineRed}{HTML}{FDF2F2}

\DeclareRobustCommand{\blue}[1]{{\setlength{\fboxsep}{1pt}\colorbox{oracleblue}{#1}}}
\DeclareRobustCommand{\cyan}[1]{{\setlength{\fboxsep}{1pt}\colorbox{BaselineRed}{#1}}}
\definecolor{improvegreen}{HTML}{0E7C3A}
\newcommand*{\improve}[1]{\textcolor{improvegreen}{+#1}}
\newcommand*{\degrade}[1]{\textcolor{crimson}{-#1}}

\newcommand{\xmark}{\ding{55}}

\newcommand{\virtualkittidataset}{VKITTI}

\newcommand{\nucarla}{nuCarla}

\newcommand{\kittidataset}{KITTI}
\newcommand{\nuscenesdataset}{nuScenes}
\newcommand{\aihubdataset}{AI-Hub}
\newcommand{\adcdataset}{ADC25}

\newcommand{\simgen}{SimGen}
\newcommand{\sodacam}{SODA-CAM}
\newcommand{\fda}{FDA}

\newcommand{\vsait}{VSAIT}
\newcommand{\sdxl}{SDXL}
\newcommand{\sdedit}{SDEdit}
\newcommand{\textimgadapter}{T2I-Adapter}
\newcommand{\controlnet}{ControlNet}
\newcommand{\dtrealsim}{DT-R2S}
\newcommand{\dtsimreal}{DT-S2R}
\newcommand{\methodname}{DT-R2S2R}

\newcommand{\fcos}{FCOS3D}
\newcommand{\detr}{DETR3D}
\newcommand{\petr}{PETR}

\newcommand{\map}{mAP}
\newcommand{\nds}{NDS}
\newcommand{\ssim}{SSIM}
\newcommand{\psnr}{PSNR}
\newcommand{\fid}{FID}
\newcommand{\lpips}{LPIPS}
\newcommand{\iou}{IoU}
\newcommand{\cornererror}{Corner Err.}

\newcommand{\realpaired}{\texorpdfstring{$D^{Real}_{InTwin}$}}
\newcommand{\realunpaired}{\texorpdfstring{$D^{Real}_{OffTwin}$}}
\newcommand{\realall}{\texorpdfstring{$D^{Real}_{All}$}}
\newcommand{\realsimpaired}{\texorpdfstring{$D^{Real2Sim}_{InTwin}$}}
\newcommand{\simrealpaired}{\texorpdfstring{$D^{Sim2Real}_{InTwin}$}}
\newcommand{\simrealadc}{\texorpdfstring{$D^{Sim2Real}_{ADC25}$}}

\newcommand{\simadc}{\texorpdfstring{$D^{Sim}_{ADC25}$}}

\newcommand{\reftable}[1]{Table~\ref{#1}}
\newcommand{\reffigure}[1]{Fig.~\ref{#1}}
\newcommand{\refsection}[1]{Section~\ref{#1}}

\newcommand{\smalltext}[1]{\,{\textcolor{mydarkgrey}{\scriptsize #1}}}

\definecolor{draftgreen}{HTML}{0B7A3B}

  {\begingroup\color{draftgreen}%
   \begin{itemize}\setlength{\itemsep}{1pt}\setlength{\parskip}{0pt}}%
  {\end{itemize}\endgroup}

\begin{document}
\ifanon
  \setlength{\dbltextfloatsep}{5pt plus 2pt minus 2pt}
  \setlength{\textfloatsep}{5pt plus 2pt minus 2pt}
  \setlength{\intextsep}{5pt plus 2pt minus 2pt}
  \setlength{\dblfloatsep}{4pt plus 2pt minus 2pt}
  \setlength{\floatsep}{4pt plus 2pt minus 2pt}
  \setlength{\abovecaptionskip}{-2pt}
  \setlength{\belowcaptionskip}{-5pt}
\else
  \setlength{\dbltextfloatsep}{5pt plus 2pt minus 2pt}
  \setlength{\textfloatsep}{5pt plus 2pt minus 2pt}
  \setlength{\intextsep}{5pt plus 2pt minus 2pt}
  \setlength{\dblfloatsep}{4pt plus 2pt minus 2pt}
  \setlength{\floatsep}{4pt plus 2pt minus 2pt}
  \setlength{\abovecaptionskip}{-2pt}
  \setlength{\belowcaptionskip}{-5pt}
\fi

\bstctlcite{IEEEexample:BSTcontrol}

\newcommand{\papertitle}{Digital Twin-Driven Real2Sim2Real: Simulator-Conditioned Generation via Paired Driving-Scene Reconstruction}
\ifanon\title{\LARGE \bf \papertitle}\else\title{\papertitle}\fi

\begin{anonblock}
\author{Anonymous Authors}
\end{anonblock}

\begin{realblock}
\author{Hojun Lim$^{*}$, Hyeongseok Jeon, Donghyun Kim, Soonyoung Jung, and Heecheol Yoo$^{*\dag}$%
\thanks{Hojun Lim is with Technical University of Munich, Germany; School of Engineering \& Design, Department of Mobility Systems Engineering, Institute of Automotive Technology and Munich Institute of Robotics and Machine Intelligence (MIRMI) (e-mail: hojun.lim@tum.de). Hyeongseok Jeon is with MORAI Inc., Seoul, South Korea (e-mail: hsjeon@morai.ai). Donghyun Kim and Soonyoung Jung are with AVP Division, Hyundai Motor Company, Gyeonggi-do, South Korea (e-mail: \{kkddh, jungsoonyoung\}@hyundai.com). Heecheol Yoo is with Seoul National University, Seoul, South Korea (e-mail: hcyoo@adr.snu.ac.kr).}%
\thanks{$^{*}$: This work was done while authors were with MORAI Inc.}%
\thanks{$^{\dag}$: Corresponding author.}%
}
\end{realblock}

\maketitle
\ifanon\thispagestyle{empty}\pagestyle{empty}\fi

\begin{abstract}
Camera-based 3D perception for autonomous driving relies heavily on large annotated datasets, and deploying such a system to a new target region typically requires data collection and annotation.
Generative augmentation has been proposed to reduce this cost, but existing approaches face a fundamental trade-off: label-conditioned methods consume the very annotations they aim to replace, while simulator-conditioned methods offer free annotations but lack visual grounding to specific real environments.
This work investigates the extent to which a digital-twin-driven Real2Sim2Real pipeline (\methodname{}) can substitute for target-region real data.
By reconstructing recorded driving clips inside a georeferenced digital twin (\dtrealsim{}), we condition a diffusion model on geometrically aligned simulator renderings, establishing a digital twin-grounded Sim2Real model (\dtsimreal{}).
As a result, DT-S2R synthesizes photorealistic driving images given low-cost yet georeferenced simulator data across both reconstructed and novel simulator scenes within digital-twin coverage.
The efficacy of generated data is verified on diverse 3D detectors. \detr{}, especially, reports 93.18\% of \map{} obtained by a target-region real-data oracle, without employing target images for detector training.
Furthermore, simple co-training with existing out-of-target real data outperforms the oracle.
Thus, \methodname{} can substantially reduce the cost of manual on-site data collection and annotation in digital twin-available districts, providing a practical foundation for scaling 3D perception. 
Comprehensive details are available at \href{https://anonymouspaperdtr2s2r.netlify.app}{\textit{https://anonymouspaperdtr2s2r.netlify.app}}.

\end{abstract}

\ifanon\else
\fi

\section{Introduction}

\begin{figure}[t]
\centering
\includegraphics[width=0.97\columnwidth]{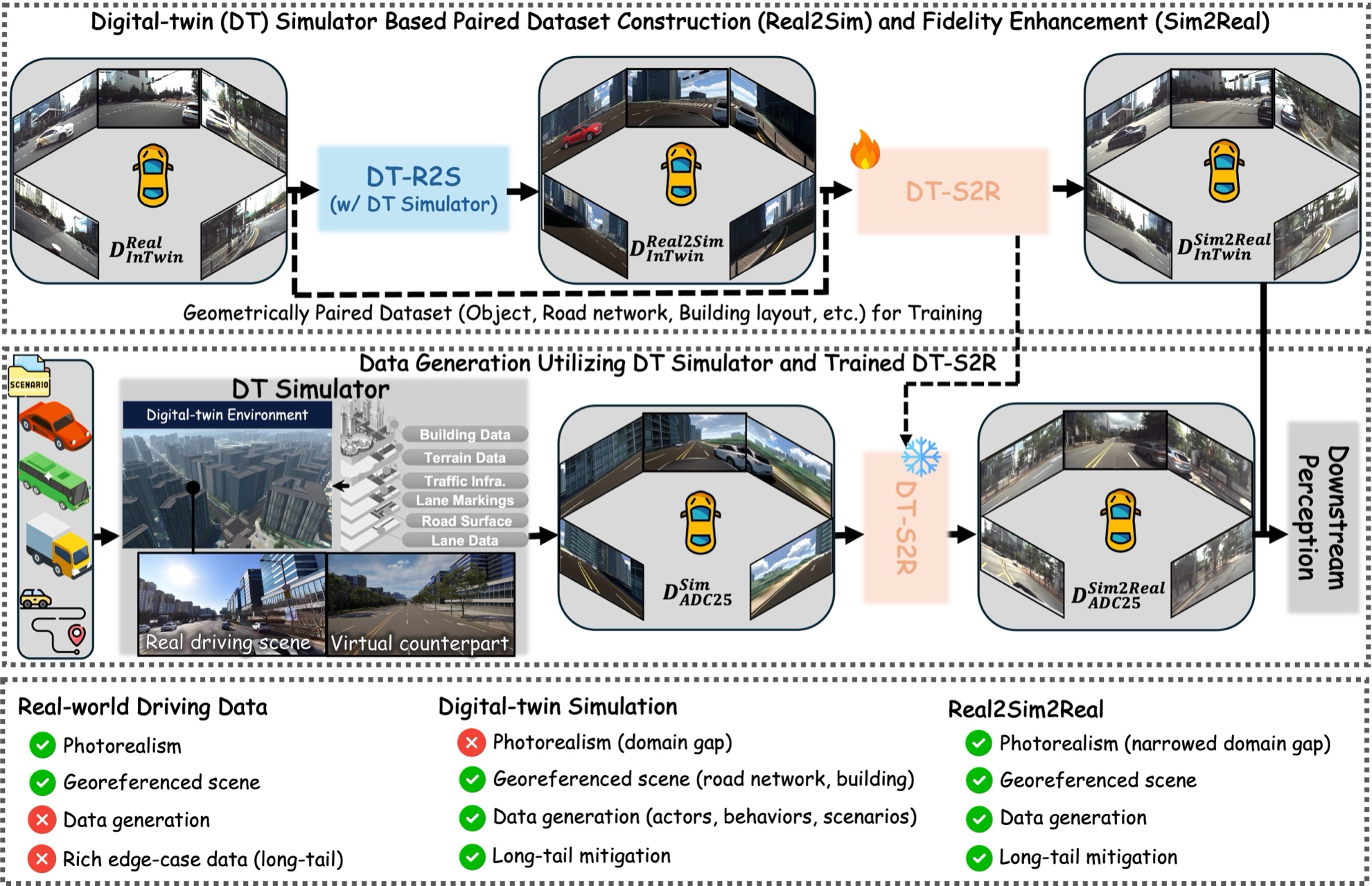}
\caption{Overview of \methodname{}. Sim2Real model is trained on real driving images and reconstruction counterparts. Then the model uses newly synthesized simulator data to produce photorealistic images for training 3D detectors.}
\label{fig_1}
\end{figure}

\parstart{C}{amera}-based 3D perception for autonomous driving is trained on large annotated datasets~\cite{nuscenes,kitti}. Despite the considerable labeling cost, it is likely to recur whenever the system needs to be deployed to an unexplored region, given that road geometry, traffic composition, vehicle fleets, and roadside architecture can all differ.
The bottleneck is thus not only the total volume of data but its regional specificity, implying that what is scarce is not driving data in general, but driving data from the region of interest.

Generative data augmentation has been proposed to relieve this cost.
One family conditions a diffusion model on the annotations of a real recording ---BEV semantic map, 3D bounding boxes, and camera poses--- and pairs each synthesized image with the annotation that produced it~\cite{gao2023magicdrive,yang2026instadriveinstanceawaredrivingworld}.
Because the generated frames inherit the geometry of an actual location, they transfer well to that location.
Generating scenes beyond the annotated set, however, requires either further manual labeling or a separate tool to produce layouts.

Driving simulators are an attractive source for these annotations.
The expensive labels that such generators require ---semantic, depth maps, and 3D bounding boxes--- are either native or easily producible outputs of the renderer and can be exported for any simulated scene at negligible marginal cost.
This motivates closing the loop: reproducing a real driving scene in a simulator (Real2Sim), exporting the conditions from that reconstruction, and generating photorealistic frames from them (Sim2Real).
Several recent Real2Sim2Real pipelines~\cite{NEURIPS2024_57c5a7c8, li2025simworld} follow this pattern.

However, in these systems, the reconstruction usually covers only the placement of dynamic objects and a coarse road layout.
The static surroundings of the location, such as building geometries and roadside structures, have no counterpart in the simulated scene and are left for the Sim2Real generator to synthesize without grounding.
To address this, we construct \textbf{D}igital-\textbf{T}win-driven \textbf{R}eal\textbf{2S}im\textbf{2R}eal (\textbf{\methodname{}}) pipeline, where the road network, buildings, terrain, and vehicle assets are modeled based on the real world rather than arbitrarily synthesized~\cite{morda, morai}.
Our pipeline consists of two modules, \dtrealsim{} and \dtsimreal{}. 
First, \dtrealsim{} locates a real driving clip inside the digital twin by matching its GNSS trace against twin maps, replaces each annotated vehicle with the library asset closest in dimension and appearance, and estimates the sensor pose aligning the reconstructed scene with the original recording. 
Through the scene rendering, we obtain synthetic RGB images geometrically paired with real counterparts and their low-cost annotations from the simulator.
\dtsimreal{} then takes them as input to condition a diffusion backbone, yielding photorealistic and geospatially grounded images. 
Importantly, producing additional virtual driving clips within the simulator and passing them to trained \dtsimreal{} enables the large-scale synthesis of high-fidelity and georeferenced data for the specific target regions at marginal cost, as shown in \reffigure{fig_1}.

The main contributions of this work are twofold. 
\begin{itemize}
    \item We construct \methodname{} that couples geometrically paired driving-scene reconstruction (\dtrealsim{}) with a spatially-consistent image translation (\dtsimreal{}).
    \item We comprehensively validate our pipeline not only as a high-fidelity image generator but also as a labeled-data engine for 3D detection training across diverse settings.  
\end{itemize}

\section{Related Work}

\begin{figure*}[b]
    \centering
    \includegraphics[width=0.9\textwidth]{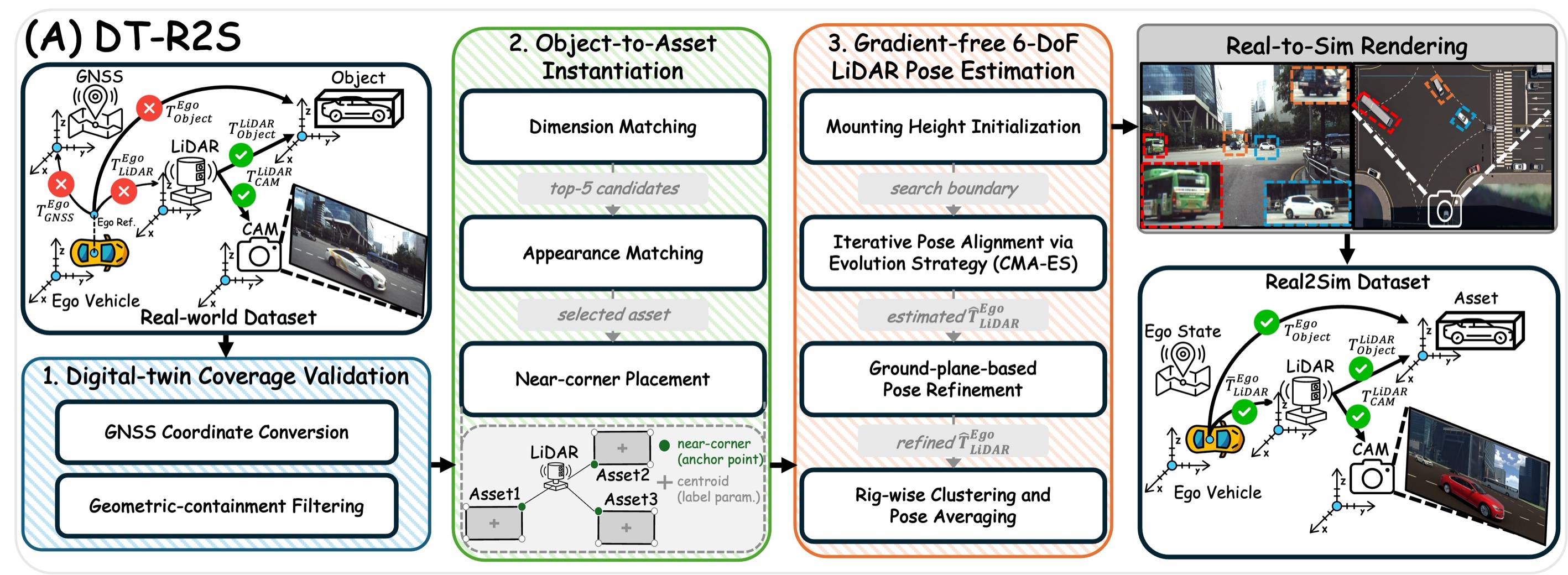}
    \caption{\textbf{Architecture of \dtrealsim{}} that locates a real driving clip inside the digital-twin simulator (1), replaces each object with the closest library asset (2), and recovers the LiDAR pose with respect to ego vehicle to improve real-sim image correspondence (3).}
    \label{fig:overview_realsim}
\end{figure*}

\subsection{Synthetic Driving Data}

Game engines and driving simulators make it affordable to produce large, labeled driving datasets, but in many cases, these datasets lack real-world counterparts.
Several datasets have addressed this by introducing direct references to real datasets to narrow the gap. For example, \nucarla{}~\cite{qiao2025nucarlanuscenesstylebirdseyeview} mirrors the sensor layout and data format of \nuscenesdataset{}~\cite{nuscenes}. 
Closer correspondence has been obtained by reconstructing the recordings. DIVA-Sim~\cite{NEURIPS2024_57c5a7c8} rebuilds \nuscenesdataset{} clips in a simulator, recovering the road network and replaying the logged object trajectories.
\virtualkittidataset{}~\cite{vkitti} goes further for \kittidataset{}~\cite{kitti}, also reproducing background structures such as buildings and trees.
Following this line, we perform Real2Sim inside a digital twin built from the HD map of the target region, where the road network, buildings, and terrain are already available at high fidelity, so that any clip recorded within the covered area can be paired with a geometrically aligned rendering.

\subsection{Photorealistic Image Generation}

The transformation of a rendered synthetic image into a realistic one has traditionally been posed as an unpaired image-to-image translation, as paired real-synthetic image sets are often not available. The characteristic limitation of those approaches is semantic flipping~\cite{vsait}. 
To overcome this limitation, \fda{}~\cite{yang2020fdafourierdomainadaptation} swaps the low-frequency amplitude of source and target images, which does not require training and retains the source contents. An adversarial learning-based method \vsait{}~\cite{vsait}, on the other hand, addresses this by learning the mapping in a high-dimensional vector space, where source-related information, such as texture and color, is unbound and target-related information is bound in its place. \sodacam{}~\cite{soda} adapts the formulation to multimodal driving data.

Another common choice for the same task is latent diffusion models (LDMs), such as \sdxl{}~\cite{ICLR2024_081b0806}, with auxiliary networks which supply conditions.
For instance, \controlnet{}~\cite{zhang2023addingconditionalcontroltexttoimage} copies the down and middle blocks of the backbone, processes the given condition, and injects their output back to the backbone.
\textimgadapter{}~\cite{mou2023t2iadapterlearningadaptersdig} replaces the copy with a small feature extractor. \sdedit{}~\cite{meng2022sdeditguidedimagesynthesis}, on the other hand, uses the noise-added condition as a starting point for the denoising process. 
Extending this direction, conditional diffusion models are also widely used to generate driving data directly from annotations. MagicDrive~\cite{gao2023magicdrive}, \simgen{}~\cite{NEURIPS2024_57c5a7c8}, and InstaDrive~\cite{yang2026instadriveinstanceawaredrivingworld} are recent examples where labels of a real recording are consumed as conditions so that the generated frames are paired with those. 

Across the mentioned methods, the output is limited by how much of the real scene the condition describes: a projected box or a semantic map states where an object is, not what it or its surroundings look like. Hence, a generator can only imagine the appearance of the surrounding structure, which often induces semantic flipping. 
We therefore use a rendering from the digital twin as a strong anchor for a paired Sim2Real translation, facilitating the preservation of critical static structures.

\section{METHODOLOGY}
\label{sec:method}

\begin{figure*}[t]
    \centering
    \includegraphics[width=0.85\textwidth]{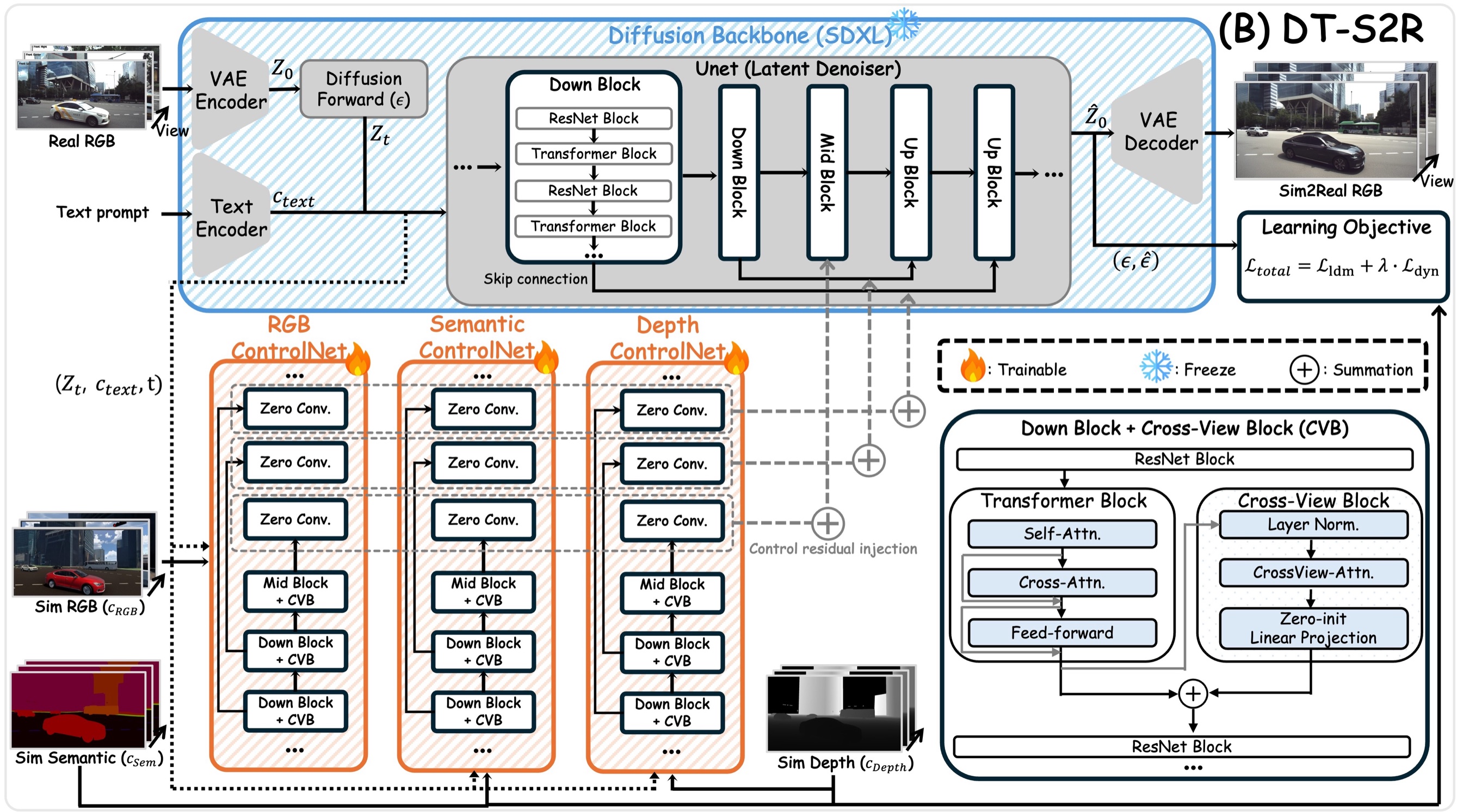}
    \caption{\textbf{Architecture of \dtsimreal{}} that drives one \controlnet{} branch per rendered modality (RGB, semantic, depth map) with cross-view blocks.}
    \label{fig:overview_simreal}
\end{figure*}

\subsection{\dtrealsim{}: Paired Driving-Scene Reconstruction}
\label{subsec:real2sim}
To generate scene-level geometrically paired counterparts, we embed real-world 
driving clips (\aihubdataset{}) of South Korea into MORAI SIM~\cite{morai}, a digital twin simulator where HD map-based road network and buildings for two urban districts, Sangam and Pangyo of the same country, together with a library of vehicle assets are available. 
As shown in \reffigure{fig:overview_realsim}, the following three steps are conducted: (1) real-to-simulator coordinate conversion, (2) real-vehicle-to-virtual-asset mapping, and (3) sensor pose estimation that closely aligns both the geometry and appearance of real images and rendered counterparts.

\subsubsection{Locating the clip in the twin}
\label{subsubsec:realsim_gnss}

We locate each recorded clip in the simulator by converting the ego vehicle's GNSS trace into the map frame via UTM projection. 
Since traces are acquired with high-accuracy RTK-DGPS, the converted trajectories align with the simulation environment without additional post-processing of the poses. 
As the digital twin covers only a finite area, we discard any clip that contains even a single out-of-bounds frame. 
To check the coverage of a clip, we extract lane-boundary nodes from the HD map and define each district's spatial extent as a convex hull. 
A clip is retained only if all of its frames fall within the hull.

\subsubsection{Replacing objects with library assets}
\label{subsubsec:realsim_asset}

After a clip is selected, we need to reproduce the dynamic objects (car, bus, and truck) of the scene inside the simulator. 
The dynamic state of each annotated object transfers directly from the recording, but two questions remain: which simulator asset substitutes for the object, and how it is positioned. 
First, we match the two by dimension and appearance. 
For box dimensions, we take the per-axis maximum over the object's tracked horizon, because a cuboid labeled from a single LiDAR sweep tends to underestimate a vehicle's true extent due to occlusion and missing returns. 
We rank the assets against the aggregated size by a weighted cost over length, width, and height. 
Each axis is weighted by its normalized coefficient of variation (length: 0.35, width: 0.24, height: 0.41), prioritizing more discriminative dimensions.
The top-5 candidates are then re-ranked by appearance: SAM~\cite{kirillov2023segment}-segmented crop of the object and each candidate asset are encoded into DINOv3~\cite{dinov3} features, and the candidate with the highest cosine similarity is kept. Second, we place assets by their near-corner positions. 
Among the four BEV corners, the one nearest to the LiDAR is the only physical measurement: annotations are drawn by fitting L- or I-shaped returns anchored on that corner, even if they are parameterized by their centroids afterward. 
We thus re-reference each object to its near corner before instantiating it in the simulator.

\subsubsection{Recovering sensor poses}
\label{subsubsec:realsim_pose}

\aihubdataset{} provides a georeferenced ego trajectory and 3D bounding boxes with respect to LiDAR frame, together with the LiDAR-to-GNSS extrinsics. However, it does not specify where the sensors are located within the vehicle body, so the sensor extrinsics with respect to the vehicle must be recovered for simulation. To resolve this, our strategy is to anchor the simulator ego's reference frame to the GNSS trajectory and deduce the remaining internal sensor placements via LiDAR data matching. 

We initialize the simulator ego's reference pose with the GNSS, transferring the horizontal placement (i.e., easting and northing in UTM) from the real data. Since the simulator grounds the vehicle to the terrain, GNSS height and tilt become redundant, leaving the LiDAR's relative height and tilt unknown. With the simulator ego anchored to the GNSS, we determine the missing LiDAR-to-ego transform ($\hat{T}_{LiDAR}^{Ego}$) by minimizing the geometric discrepancy between real and simulated bounding boxes within the LiDAR frame:
\begin{equation}
\small
\label{eq:box_corner_matching}
\hat{T}_{LiDAR}^{Ego} \;=\; \arg\min_{T}
\;\frac{1}{|\mathcal{M}(T)|}
\sum_{(i,i') \in \mathcal{M}(T)} \sum_{j=1}^{8}
\left\|
  c_{ij}^{\mathrm{real}} \;-\; c_{i'j}^{\mathrm{sim}}(T)
\right\|^{2},
\end{equation}%
where $c_{ij}^{\mathrm{real}}$ is the $j$-th corner of the $i$-th real box in the LiDAR frame (fixed, from the dataset), $c_{i'j}^{\mathrm{sim}}(T)$
is the corresponding corner of the simulated box $i'$, and $\mathcal{M}(T)$ denotes the real-to-simulated box correspondence. To optimize this objective, we set the initial condition of $T$ using the provided LiDAR-to-GNSS extrinsics. Furthermore, to define a plausible search boundary for the optimizer, we first fit a ground plane to the real point cloud via RANSAC. This provides a rough estimate of the LiDAR's physical mounting height relative to the ground, establishing a valid bounding margin for the search space. Starting from this configuration, we iteratively optimize the full 6-DoF pose of $T$ using CMA-ES~\cite{hansen2019pycma}. At each step, while the simulator ego remains fixed, updating the candidate LiDAR extrinsics repositions the surrounding vehicles within the simulator. This allows us to calculate the loss via Equation~\ref{eq:box_corner_matching}, driving CMA-ES to propose the next LiDAR pose until convergence.

However, this optimization can still leave residual misalignments due to discrepancies in vehicle dimensions and terrain elevation. To address this, we apply a final ground-plane-based refinement. Following the LiDAR-to-ego calibration procedure of~\cite{complexurban}, we fit a 
ground plane to both the simulated and real LiDAR point clouds independently and use the extracted roll, pitch, and $z$ values to correct the remaining LiDAR pose error. Finally, analyzing the calibrations of \aihubdataset{} reveals four distinct sensor rig setups. We group the per-clip estimates $\hat{T}_{LiDAR}^{Ego}$ accordingly, average translations arithmetically and rotations via SLERP, and adopt the mean as the stable $\bar{T}_{LiDAR}^{Ego}$ for each rig. 
\ifanon\else\enlargethispage{0.5\baselineskip}\fi
\begin{table*}[b]
\caption{Summary of individual datasets and their characteristics. ($\#$Clip, $\#$Frame) is noted in Train and Val. columns. DT Cover. denotes the digital twin coverage. \realall{} denotes the combination of \realpaired{} and \realunpaired{}.}
   \label{tab:dataset_properties}
   \centering
\makebox[\columnwidth][c]{%
    \resizebox{0.8\textwidth}{!}{%
      \begin{threeparttable}[b]
       \begin{tabular}{llllllll}
        \toprule[1.5pt]
          \bfseries{Dataset} & \bfseries{Domain} & \bfseries{DT Cover.} & \bfseries{\dtsimreal{}} & \bfseries{Region/District} & \bfseries{Train} & \bfseries{Val.} & \bfseries{Notation} \\
         \midrule
            \vspace{2pt}
            \aihubdataset{} (Off-Twin) & Real & \xmark{} & N/A & Seoul, Seongnam, Sejong, Daegu & (262, 47K) & (66, 11K) & \realunpaired{} \\ 
            \vspace{2pt}
            \aihubdataset{} (In-Twin) & Real & \pmb{\checkmark{}} & N/A & Sangam, Pangyo & (134, 24K) & (34, 6K) & \realpaired{} \\ 
            \vspace{2pt}
            \dtrealsim{} & Synthetic & \pmb{\checkmark{}} & \xmark{} & Sangam, Pangyo & (134, 24K) & (34, 6K) & \realsimpaired{} \\ 
            \vspace{2pt}
            \dtsimreal{} & Synthetic & \pmb{\checkmark{}} & \pmb{\checkmark{}} & Sangam, Pangyo & (134, 24K) & (34, 6K) & \simrealpaired{} \\ 
         \midrule
            \vspace{2pt}
            \adcdataset{} & Synthetic & \pmb{\checkmark{}} & \xmark{} & Sangam, Pangyo & (241, 28K) & (0, 0) & \simadc{} \\ 
            \vspace{2pt}
            \adcdataset{} (\dtsimreal{}) & Synthetic & \pmb{\checkmark{}} & \pmb{\checkmark{}} & Sangam, Pangyo & (241, 28K) & (0, 0) & \simrealadc{} \\ 
        \bottomrule[1.5pt]
         \end{tabular}
      \end{threeparttable}
    }%
}
\end{table*}
\subsection{\dtsimreal{}: Photorealism Enhancement}
\label{subsec:sim2real}
Reconstruction in a digital twin paired with the real-world recording already provides a wealth of information for Sim2Real generator to leverage: a robust anchor for the placement of dynamic objects and a reference for static structures that every rendered view coherently shares. 
In this sense, we hypothesize that minimal modifications to the generative network could be sufficient. 
Consequently, the backbone is frozen throughout, and our modifications are focused on how the conditions are injected.

We take three conditions from the reconstruction: depth $c^{\mathrm{dep}}$, semantics $c^{\mathrm{sem}}$, and the simulator's RGB view $c^{\mathrm{rgb}}$. Those three are selected as they convey complementary information. Depth and semantics fix \emph{where} surfaces are and \emph{what} they are. On the other hand, the rendered view additionally fixes \emph{what the place and objects look like}, since the road network, the surrounding buildings and the substituted assets all appear in it.
Each condition $k$ drives a ControlNet $\mathcal{C}_k$, a trainable copy of the down and middle blocks of the backbone~\cite{zhang2023addingconditionalcontroltexttoimage}. Each ControlNet returns one residual per skip connection, together with one for the middle block,
\begin{equation}
\big(r^{k}_{1},\dots,r^{k}_{L},\; r^{k}_{\mathrm{mid}}\big)
  = \mathcal{C}_k\!\left(z_t,\, t,\, c^{\mathrm{text}},\, c^{k}\right),
\label{eq:branch}
\end{equation}
where $z_t$ is the noisy latent, $t$ the timestep and $c^{\mathrm{text}}$ the text embedding.
Writing $h_{\ell}$ for the backbone's own $\ell$-th skip feature and $h_{\mathrm{mid}}$ for its middle-block output, the enriched features can be denoted as
\begin{equation}
\tilde h_{\ell}=h_{\ell}+\sum_{k} s_k\, r^{k}_{\ell},
\qquad
\tilde h_{\mathrm{mid}}=h_{\mathrm{mid}}+\sum_{k} s_k\, r^{k}_{\mathrm{mid}},
\label{eq:inject}
\end{equation}
with $s_k=1$ for all three.
As \eqref{eq:inject} formulates, the integration mechanism is conducted not by modifying the diffusion backbone directly. 
Instead, the three simulator conditions influence the denoising process via additive residual features injected into the backbone's skip and middle blocks.

Let $\Omega$ be the set of spatial locations in the latent grid and $\Omega_{\mathrm{dyn}}\subset\Omega$ be those belonging to dynamic objects.
Then we optimize the following loss: 
\begin{equation}
\mathcal{L}_{total}=\frac{1}{|\Omega|}\sum_{u\in\Omega}\big\lVert\hat\epsilon_u-\epsilon_u\big\rVert^{2}
\;+\;\frac{\lambda}{|\Omega_{\mathrm{dyn}}|}\sum_{u\in\Omega_{\mathrm{dyn}}} w_u\big\lVert\hat\epsilon_u-\epsilon_u\big\rVert^{2},
\label{eq:loss}
\end{equation}
where $\hat\epsilon$ is the predicted noise, and $w_u=\max(d_u,\,0.02)$ grows with the normalized depth $d_u$ from $c^{dep}$.
The first term is the learning objective of LDM. The second term, inspired by the reweighting mechanism of~\cite{lu2024cogdriving}, exists because distant objects occupy fewer pixels but convey most of the difficulty for 3D detection, and an error averaged over all spatial positions would otherwise ignore them.

In light of the multiview-camera setup being pivotal in 3D perception, we additionally equip our method with multiview-image generation by employing the cross-view blocks following~\cite{gao2023magicdrive}. 
Specifically, for each camera $i$, the hidden state $a_i$ gets updated by taking cross attention with that of left and right neighbors $\mathcal{N}(i)$: 
\begin{equation}
a_i \leftarrow a_i + W_0\!\left(\sum_{j\in\mathcal{N}(i)}\mathrm{Attn}\big(\mathrm{LN}(a_i),\,\mathrm{LN}(a_j)\big)\right),
\label{eq:cva}
\end{equation}
with layer normalization $\mathrm{LN}$ and zero-initialized $W_0$ so that the term vanishes at initialization.
As depicted in \reffigure{fig:overview_simreal}, we insert the cross-view blocks into the conditioning branches.

\section{EXPERIMENTAL SETUP}
\label{sec:experimental_setup}
\subsection{Datasets}

As summarized in \reftable{tab:dataset_properties}, we use one real and two synthetic datasets to demonstrate our approach. \aihubdataset{} is a real driving dataset recorded on urban roads of South Korea with five surround-view cameras and one 128-channel LiDAR.
We divide it by regional coverage of a digital-twin simulator: a clip is in-twin (\realpaired{}) if its GNSS trajectory lies inside the districts of digital-twin maps so that the virtual counterparts exist, and off-twin (\realunpaired{}) otherwise. \realall{} denotes the two together. 
Note that the official training and validation splits are frame-wise. Hence, we repartition \aihubdataset{} driving clips in clip-wise manner for fair training of Sim2Real generator, fairly distributing dynamic objects in polar coordinate system with respect to the ego vehicle across splits.

Applying \dtrealsim{} to \realpaired{} creates \realsimpaired{}, the rendered counterpart of each recording, and passing that to \dtsimreal{} gives \simrealpaired{}.
\adcdataset{} is a third dataset that did not participate in constructing the pipeline and is used solely to test the scalability of our trained \dtsimreal{} on pure simulator-generated driving clips.
Its scenarios originate from the dataset of Hyundai Motor Group Autonomous Driving Challenge 2025. We only select scenarios whose trajectories lie within the digital-twin coverage as \realpaired{}. The scenarios are re-simulated with the reconstructed sensor pose, giving \simadc{}. 
\dtsimreal{} then produces \simrealadc{}. Note that no frame of \simrealadc{} corresponds to a real recording, which means producing it did not require real-world data collection nor new labeling, but rather re-using the digital-twin simulator and trained \dtsimreal{}.

\begin{figure*}[b]
    \centering
    \includegraphics[width=0.95\textwidth]{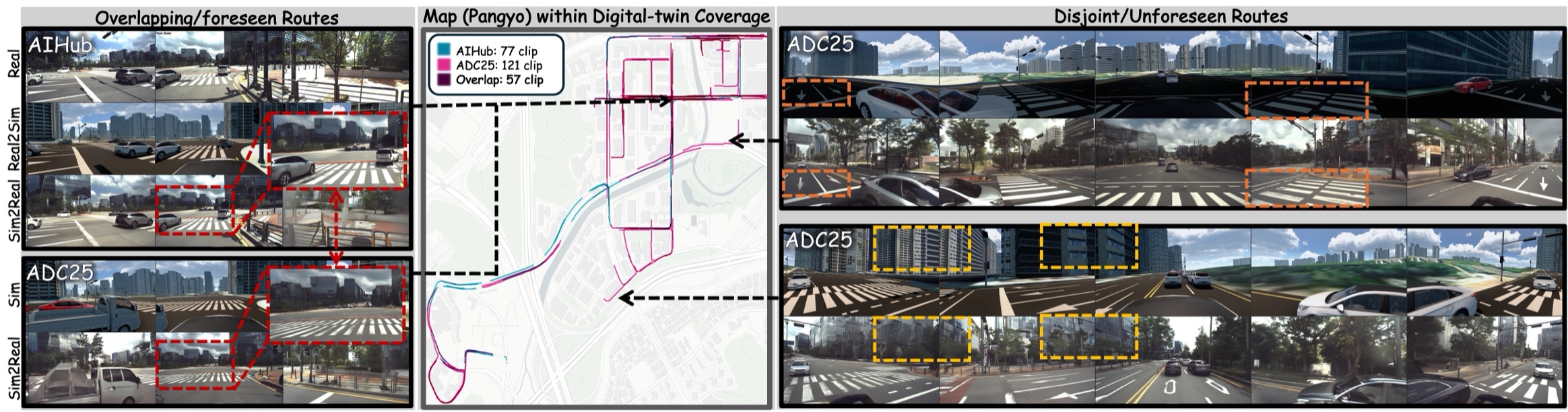}
    \caption{Data generation using \dtsimreal{} conditioned on both known and unknown driving routes from the simulator. (\romannum{1}) Geographical features and buildings are retained where the digital twin supplies them. (\romannum{2}) Road-network detail survives \dtsimreal{}. (\romannum{3}) Cross-view consistency holds on a route absent from training.}
    \label{fig:adc_overlap_nooverlap}
\end{figure*}

While both \realpaired{} and \simrealadc{} cover the same district, their street-level routes diverge. In Pangyo, for example, $\sim$53\% clips of \simrealadc{} are disjoint as visualized in \reffigure{fig:adc_overlap_nooverlap}.
These clips allow us to visually inspect whether our \dtsimreal{} can synthesize image data across unvisited streets, leveraging the rendered scene within the digital-twin area.

\subsection{Implementation details}
\label{subsec:training_config}

\subsubsection{\dtsimreal{} training}
All views are resized to $576\times1024$.
The diffusion backbone, its VAE, and both text encoders are frozen, and only the three conditioning branches are trained.
We use AdamW with learning rate $5\times10^{-5}$, weight decay $0.01$, $1000$ warm-up steps, and a per-device batch of two frames $\times$ five cameras on eight NVIDIA RTX A6000 GPUs.
Training proceeds in two stages.
Stage~1 trains the conditioning branches without cross-view blocks for $50{,}000$ steps with $\lambda$ of \eqref{eq:loss} set to 0.1.
Stage~2 initializes from the stage-1 checkpoint, enables cross-view blocks in the \texttt{mid} and the two outermost \texttt{down} blocks of each branch, raises $\lambda$ to $0.3$, and trains for a further $40{,}000$ steps. 
The prompt is set to \texttt{highly detailed realistic driving scene in \{district\}, South Korea}, and it is dropped with probability $0.1$.
Lastly, the sampling process is performed with DDPM scheduler at $50$ steps and guidance scale $5.0$.

\subsubsection{Other design choices for \dtsimreal{}}
The set of image-to-image translators ---\fda{}~\cite{yang2020fdafourierdomainadaptation}, \vsait{}~\cite{vsait} and \sodacam{}~\cite{soda}--- takes the pairs of real and synthetic RGB images of \realpaired{} and \simrealpaired{}, following the training configuration of~\cite{soda}. 
\simgen{} (ImgDiff)~\cite{NEURIPS2024_57c5a7c8} is trained taking RGB, semantic and depth labels of \realsimpaired{} as conditions. 

The synthetic RGB rendered using the digital-twin simulator is the strongest of our three conditions, and while designing the conditioning path, we considered \sdedit{}~\cite{meng2022sdeditguidedimagesynthesis} and \textimgadapter{}~\cite{mou2023t2iadapterlearningadaptersdig} as two other ways of injecting it.
Both methods are implemented based on stage-1 configuration.
\sdedit{} places the synthetic RGB in the initial latent at $t_0=0.5$ and leaves depth and semantics as \controlnet{} branches.
 On the other hand, \textimgadapter{} passes all three signals through lightweight adapters with a smaller learning rate $1\times10^{-5}$. 

\subsubsection{3D Detectors}
Downstream perception is measured with one monocular 3D detector, \fcos{}~\cite{wang2021fcos3d}, two multi-view 3D detectors, \detr{}~\cite{detr3d} and \petr{}~\cite{liu2022petr}.

\subsection{Metrics}

The quality of generated images is measured with four complementary quantities.
\psnr{} and \ssim{} quantify pixel-level and structural agreement with the paired real frame. \lpips{} measures distance to real image at deep-feature level. Lastly, \fid{} compares the real and generated image sets at the distribution level. For downstream perception tasks, \map{} and \nds{}~\cite{nuscenes} across car, bus, and truck classes are measured for 3D detection. Scores are evaluated on a \realpaired{} validation set unless otherwise mentioned.

\section{Experiments}

We first validate \dtrealsim{} and \dtsimreal{} separately in \refsection{subsec:design}.
Then, \refsection{subsec:proxy} asks whether the generated data can replace real counterparts for 3D detector training, and \refsection{sec:enriching_diversity} studies the extent to which performance can be achieved without real target-region images by exploiting the digital-twin simulation and \methodname{} for the same region.

\subsection{Design Choices for \methodname{}}
\label{subsec:design}

\subsubsection{\dtrealsim{}: scene reconstruction and sensor alignment}
Table~\ref{tab:real2sim_ablation} isolates each stage of \refsection{subsubsec:realsim_asset} and \ref{subsubsec:realsim_pose} on the raw reconstruction. 
2D IoU is the image-space IoU between real and simulated cuboids, each projected onto the camera in which the simulated box is largest. 
3D Corner Error is the mean distance over all corners of real and asset 3D boxes. 

The baseline---per-axis-max dimension aggregation with near-corner placement and the box-corner pose estimate---reaches 57.5 IoU and 0.581\,m corner error. Appearance re-ranking trades a marginal amount of geometric fit for visual similarity (56.7, 0.632\,m). 
This could be explained as the visually closest asset need not be the dimensionally closest one among candidates with near-tied dimensions, and corner error conflates that mismatch with pose error. 
The two remaining stages recover and surpass the baseline on all metrics. Ground-plane refinement raises IoU to 59.4. Clustering per-clip estimates into one transform per calibration group then gives the best scores throughout (60.4, 0.565\,m), indicating the importance of a proper strategy to address per-clip pose noise.
Nevertheless, precision remains bounded by the finite asset catalog, underscoring the need for automated asset reconstruction from sensor data as a key enabler of scaling.

\begin{table}[tb]
\centering
\caption{Ablation study on \dtrealsim{}. Best scores are shown in \textbf{bold}, and second-best scores are
\underline{underlined}.}
\label{tab:real2sim_ablation}
\small 
\resizebox{0.95\columnwidth}{!}{%
    \begin{tabular}{lcccc}
    \toprule
        \textbf{Ablation} & \textbf{2D \iou{}}$\uparrow$ & \textbf{3D \cornererror{}}$\downarrow$ & \textbf{\ssim{}}$\uparrow$ & \textbf{\lpips{}}$\downarrow$  \\
        \midrule
        Baseline & 57.5\% & \underline{0.581} & 0.3427 & 0.6836 \\
        + Appearance re-ranking & 56.7\% & 0.632 & 0.3420 & 0.6849\\
        + Ground-plane refinement & \underline{59.4\%} & 0.582 & \underline{0.3430} & \underline{0.6809} \\
        + Clustering \& Averaging & \textbf{60.4\%} & \textbf{0.565} & \textbf{0.3458} & \textbf{0.6804} \\
        \bottomrule
    \end{tabular}
    }
\end{table}

\subsubsection{\dtsimreal{}: how the rendered view is used for paired Sim2Real training}
Despite the sophisticated design for geometrically paired reconstruction, there exists a non-negligible visual domain gap to the real counterparts, coming from the limitation of the rendering pipeline. To address this, we additionally employ Sim2Real translation and evaluate it across three families --- signal-processing, GAN, and diffusion. As presented in \reftable{tab:sim2real_image_quality}, diffusion-based methods generally produce more realistic results than other methods. Based on the observation, we develop our \dtsimreal{} (Stage1) upon the frozen diffusion backbone with ControlNet. For comparison, we also vary the injection mechanism of rendered RGB across two configurations: \sdedit{} and \textimgadapter{}. 

\begin{figure}[!b]
\centering
\includegraphics[width=1\columnwidth]{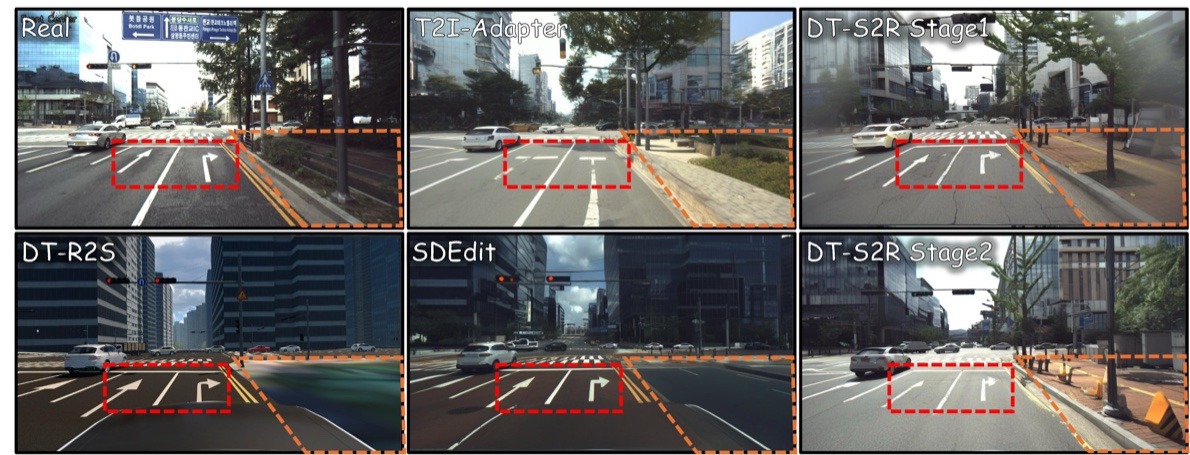}
\caption{Visual comparison of translated images by injection mechanisms of rendered-RGB at Sim2Real step.}
\label{fig:qual_comparison}
\end{figure}

\sdedit{} injects it as an initial latent, reaching 57.2 \fid{}. \textimgadapter{}, on the other hand, passes it through a lightweight adapter, presenting 22.1. Lastly, \dtsimreal{} (Stage1) that uses residual conditioning branches for RGB reports 13.5, the better score over compared mechanisms, and the same holds for \psnr{}, \ssim{}, and \lpips{}. 
\reffigure{fig:qual_comparison} shows where each mechanism differs from the target. 
\sdedit{} holds the road network and the geometry of the conditioning render closely, but the appearance of the simulator stays visible in its output.
\textimgadapter{} keeps the global layout but loses thin structures such as lane markings.
Further training Stage1 with cross-view blocks leads to the proposed \dtsimreal{} (Stage2), reporting the best structural similarity and fidelity on each metric in table.

It is worth noting that the behavior of \sdedit{} is one end of a trade-off.
Because it alters the conditioning RGB conservatively, the rendered structure remains after translation, leading to a lower chance of semantic flipping. This characteristic would make \sdedit{} more suitable as a training data engine for dense perception tasks such as segmentation, where pixel-level alignment between the generated labels and translated images is essential. 

\begin{table}[tb]
\centering
\caption{Image quality in terms of similarity to paired validation set \realpaired{} by Sim2Real methods.}
\label{tab:sim2real_image_quality}
\small
\begin{threeparttable}
\makebox[\columnwidth][c]{%
    \resizebox{0.8\columnwidth}{!}{%
        \begin{tabular}{lcccc}
        \toprule
            \textbf{Model} & \textbf{\psnr{}}$\uparrow$ & \textbf{\ssim{}}$\uparrow$ & \textbf{\lpips{}}$\downarrow$ & \textbf{\fid{}}$\downarrow$ \\
            \midrule
            \multicolumn{5}{l}{\textit{\textbf{Real2Sim}}} \\
            \dtsimreal{} (\realsimpaired{})  & 9.062 & 0.354 & 0.676 & 101.3 \\
            \midrule
            \multicolumn{5}{l}{\textit{\textbf{Signal-Processing based Sim2Real}}} \\
            \fda{}\textsuperscript{\ensuremath{\dagger}} \cite{yang2020fdafourierdomainadaptation} & 10.168 & 0.414 & 0.682 & 117.8 \\
            \midrule
            \multicolumn{5}{l}{\textit{\textbf{GAN-based Sim2Real}}} \\
            \vsait{} \cite{vsait} & 9.478 & 0.391 & 0.648 & 85.1 \\
            \sodacam{} \cite{soda} & 10.094 & 0.409 & 0.649 & 93.5 \\
            \midrule
            \multicolumn{5}{l}{\textit{\textbf{Diffusion-based Sim2Real}}} \\
            \simgen{}\textsuperscript{\ensuremath{\ddagger}} \cite{NEURIPS2024_57c5a7c8} & \underline{12.040} & 0.440 & 0.527 & 21.3 \\
            \sdedit{}\textsuperscript{\ensuremath{\dagger}} \cite{meng2022sdeditguidedimagesynthesis} & 9.458 & 0.397 & 0.650 & 57.2 \\
            \textimgadapter{}\textsuperscript{\ensuremath{\dagger}} \cite{mou2023t2iadapterlearningadaptersdig} & 9.890 & 0.367 & 0.634 & 22.1 \\
            \dtsimreal{}-Stage1 (Ours) & 11.778 & \underline{0.452} & \underline{0.509} & \underline{13.5} \\
            \rowcolor{tablegrey} \dtsimreal{}-Stage2 (Ours) & \textbf{12.104} & \textbf{0.461} & \textbf{0.493} & \textbf{12.7} \\
            \bottomrule
        \end{tabular}%
        }
    }
    \begin{tablenotes}[flushleft]
        \footnotesize
        \item[] $\dagger$: Re-implemented. $\ddagger$: Only the ImgDiff module of \cite{NEURIPS2024_57c5a7c8} is used.
    \end{tablenotes}
\end{threeparttable}
\end{table}

\subsubsection{Do real-world-trained detectors read the generated frames as they read real ones?}

Image-quality metrics state how similar two images look, not whether a perception model reads them the same way.
We therefore comprehensively run one mono-view and two multiview detectors trained on \aihubdataset{} (\realpaired{}) over \dtrealsim{} (\realsimpaired{}) and \dtsimreal{} (\simrealpaired{}) outputs, respectively. 
We then score their predictions against the original real ground truth in \reftable{tab:detector_agreement}.
Note that the scores show the disagreement accumulating two error sources, the geometry misalignment from the scene-recovery step and appearance gap from the translation step.

The reconstruction on its own already recovers 47.5--57.6\% of the \map{} that the same detectors obtain on the corresponding real frames.
The Sim2Real step further raises \map{} to near 70\% for \detr{} and \nds{} over 76\% for all detectors, demonstrating the fidelity of data generated by the entire pipeline \methodname{} against original real scenes.

\subsection{Generated Data as a Valid Proxy for Real Training Data}
\label{subsec:proxy}

\begin{table}[tb]
\centering
\caption{Image fidelity quantified by \fcos{} trained on \realpaired{}. For validation clips of \realpaired{}, it takes each of the three rows as an input. The predictions by rows are scored against ground truth. Row colored \blue{blue} denotes oracle.}
\label{tab:detector_agreement}
\small
\begingroup
\setlength{\tabcolsep}{2.5pt}
\resizebox{0.9\columnwidth}{!}{%
    \begin{tabular}{@{}l@{\hspace{3pt}}cc@{\hspace{4pt}}cc@{\hspace{4pt}}cc@{}}
    \toprule
        \multirow{3}{*}{\shortstack[c]{\textbf{Inf. Input} \\ \textbf{(val. set)}}} & \multicolumn{2}{c}{\textbf{Mono-view}} & \multicolumn{4}{c}{\textbf{Multi-view}} \\
        \cmidrule(lr){2-3} \cmidrule(lr){4-7}
        & \multicolumn{2}{c}{\textbf{\fcos{}}} & \multicolumn{2}{c}{\textbf{\detr{}}} & \multicolumn{2}{c}{\textbf{\petr{}}} \\
        \cmidrule(lr){2-3} \cmidrule(lr){4-5} \cmidrule(lr){6-7}
        & \textbf{\map{}}$\uparrow$ & \textbf{\nds{}}$\uparrow$ & \textbf{\map{}}$\uparrow$ & \textbf{\nds{}}$\uparrow$ & \textbf{\map{}}$\uparrow$ & \textbf{\nds{}}$\uparrow$ \\
        \midrule
            \rowcolor{oracleblue} \vspace{2pt}
            \realpaired{} & 52.2 & 57.4 & 53.4 & 57.6 & 60.4 & 61.0 \\
        \midrule
            \vspace{2pt}
            \realsimpaired{} & 24.8\smalltext{(47.5\%)} & 40.0\smalltext{(69.6\%)} & 30.8\smalltext{(57.6\%)} & 42.3\smalltext{(73.4\%)} & 31.1\smalltext{(51.4\%)} & 41.5\smalltext{(68.0\%)} \\
            \rowcolor{tablegrey} \vspace{2pt}
            \simrealpaired{} & 32.0\smalltext{(61.3\%)} & 45.6\smalltext{(79.4\%)} & 37.0\smalltext{(69.2\%)} & 47.5\smalltext{(82.4\%)} & 35.5\smalltext{(58.7\%)} & 46.8\smalltext{(76.7\%)} \\
        \bottomrule
    \end{tabular}
    }
\endgroup
\end{table}

\subsubsection{Is the Sim2Real step necessary?}
We first explore whether the Sim2Real step brings performance gains to 3D detector training.
\reftable{tab:necessity_and_augmentation}(a) shows the performance of \fcos{} trained solely on the synthetic data with and without \dtsimreal{}. 
The same scenes after the translation step yield a higher \map{} by nearly a factor of 5 than raw reconstruction, indicating that reproducing geometries in the simulator alone does not produce trainable data due to the large visual domain gap.

\subsubsection{How far does the generated data substitute for real data of the same scene?}
In \reftable{tab:substitution_sweep} we vary the ratio of real to the paired synthetic data, holding the total number of training frames fixed during \fcos{} training. 
As intuitively expected, in general, the performance declines gradually as real data are withdrawn on both \map{} and \nds{}. 
However, even when all real data are retracted, the detector retains 90.28\% and 93.23\% of its oracle \map{} and \nds{}, respectively. 
This demonstrates that the high-quality reproduction of real-world scenes by \methodname{} yields valid synthetic proxies, serving as a viable alternative to real-world data for 3D detector training.

\subsubsection{Does the generated data also augment?}
As shown in \reftable{tab:necessity_and_augmentation}(b), adding the generated copies to the full real data raises \map{} from 37.59 to 38.60, even though the two sets depict fundamentally the same scenes. This is attributed to \dtsimreal{} which maintains the dynamic-object placement but resynthesizes the color, texture, and lighting. In other words, it supplies the variation that the real data do not contain.
\begin{table}[t]
\centering
\caption{Detection performance of \fcos{}-\texttt{w/o finetune} trained for 12 epochs and evaluated on \realall{}.}
\ifanon\vspace{-1pt}\else\vspace{-8pt}\fi
\label{tab:necessity_and_augmentation}
\small
\hfill
\begin{minipage}[t]{0.21\textwidth}
\vspace{0pt}
\centering
\parbox[t][2.5em][c]{\linewidth}{\centering (a) Effect of \dtsimreal{} when training on synthetic data.}\\[2pt]
\vspace{2pt}
\resizebox{\linewidth}{!}{%
\begin{tabular}{lcc}
\toprule
\textbf{Training set} & \textbf{\map{}}$\uparrow$ & \textbf{\nds{}}$\uparrow$ \\
\midrule
\realsimpaired{} & 6.86 & 27.31 \\
\rowcolor{tablegrey} \vspace{1pt}\simrealpaired{} & \textbf{33.94} & \textbf{45.93} \\
\bottomrule
\end{tabular}%
}
\end{minipage}
\begin{minipage}[t]{0.27\textwidth}
\vspace{0pt}
\centering
\parbox[t][2.8em][c]{\linewidth}{\centering (b) Effect of generated data in augmentation. The number of epochs is halved when w/ \simrealpaired{}.}\\[2pt]
\vspace{1pt}
\resizebox{\linewidth}{!}{%
\begin{tabular}{lcc}
\toprule
\textbf{Training set} & \textbf{\map{}}$\uparrow$ & \textbf{\nds{}}$\uparrow$ \\
\midrule
\realpaired{} & 37.59 & 49.26 \\
\rowcolor{tablegrey} \realpaired{} + \simrealpaired{} & \textbf{38.60} & \textbf{50.08} \\
\bottomrule
\end{tabular}%
}
\end{minipage}
\end{table}

\begin{table}[tb]
\centering
\caption{Replacement of real data with generated counterparts. The experiment setup is the same as \reftable{tab:necessity_and_augmentation}. Oracle is in \blue{blue}.}
\label{tab:substitution_sweep}
\small 
\makebox[\columnwidth][c]{%
    \resizebox{0.85\columnwidth}{!}{%
        \begin{tabular}{lll|cc}
        \toprule
            \vspace{2pt}
            \textbf{Real} & \textbf{Synthetic} & \textbf{Ratio} & \textbf{\map{}}(\%, $\uparrow$) & \textbf{\nds{}}$(\%, \uparrow$) \\
        \midrule
            \addlinespace[2pt]
            \rowcolor{oracleblue} \realpaired{} & \xmark{} & 100:0 & 37.59\smalltext{(100.00\%)} & 49.26\smalltext{(100.00\%)} \\
        \midrule
            \vspace{2pt}
            \realpaired{} & \simrealpaired{} & 90:10 & 37.67\smalltext{(100.21\%)} & 49.22\smalltext{(99.91\%)} \\
            \vspace{2pt}
            \realpaired{} & \simrealpaired{} & 75:25 & 36.68\smalltext{(97.57\%)} & 48.61\smalltext{(98.68\%)} \\
            \vspace{2pt}
            \realpaired{} & \simrealpaired{} & 50:50 & 36.13\smalltext{(96.11\%)} & 48.06\smalltext{(97.56\%)} \\
            \vspace{2pt}
            \realpaired{} & \simrealpaired{} & 25:75 & 34.84\smalltext{(92.68\%)} & 47.37\smalltext{(96.16\%)} \\
            \vspace{2pt}
            \realpaired{} & \simrealpaired{} & 10:90 & 34.30\smalltext{(91.24\%)} & 47.04\smalltext{(95.49\%)} \\
            \rowcolor{tablegrey} \vspace{2pt} \xmark{} & \simrealpaired{} & 0:100 & 33.94\smalltext{(90.28\%)} & 45.93\smalltext{(93.23\%)} \\
        \bottomrule
        \end{tabular}
        }%
    }
\end{table}

\subsection{Enriching Scene-Level Diversity with Novel Synthetic Sim2Real Data}
\label{sec:enriching_diversity}

\subsubsection{How Close Can Expanded Scene Diversity Bring Us to the Real-Data Oracle?}
In this section, we evaluate the feasibility of deploying a 3D detection system to the target region without directly utilizing any target-region real images for detector training. 
To this end, we investigate the effect of combining two synthetic datasets: reconstructed clips (\simrealpaired{}) and \adcdataset{} (\simrealadc{}), a purely simulator-generated dataset using the digital twin and trained \dtsimreal{}. 

\reftable{tab:region_without_real_3d} summarizes the efficacy of our generated data. Training on \simrealadc{} alone achieves 29.39 and 34.41 \map{} on \detr{} and \petr{}, which are close to 61\% of oracle performance. Additionally, when combined with \simrealpaired{}, \map{} of \detr{} increases further to 45.50, 93.18\% of oracle score, and the same trend can be observed for \petr{}, reporting 88.84\% of oracle performance. Importantly, when compared against the practical real-world baseline trained on out-of-target region data (\realunpaired{}), it outperforms the baseline by +5.52 \map{} on \detr{} and achieves a comparable score on \petr{}. Although the non-negligible performance gaps of 3.33 \map{} for \detr{} and 6.16 \map{} for \petr{} to the oracle scores still remain, these results show that our digital twin-grounded synthetic data can effectively substitute for real-world data collected and labeled for the target region.

\subsubsection{Can Off-Target Real Data Complement Synthetic Diversity to Reach Oracle Performance?}
To bridge the remaining gap, we additionally incorporate out-of-target real data into our full synthetic datasets. 
This joint training achieves 51.02 \map{} on \detr{} and 55.79 \map{} on \petr{}, exceeding the target-region real oracle by +2.19 \map{} and +0.56 \map{}, respectively.

In a distance-based breakdown analysis, the practical value of this joint training can be clearly observed. 
In a mid-to-long range (30--50m) where stable detection is pivotal for robust downstream prediction and planning, the practical baseline trained on \realunpaired{} yields 23.56 (78.9\% oracle) on \detr{} and 32.31 \map{} (85.16\% oracle) on \petr{}. 
Crucially, these notable gaps to the oracle are alleviated when also employing our full synthetic suite. Especially for \detr{}, the \map{} gain of +9.99 is observed, surpassing the oracle by +3.69.

\subsection{Long-Tail Scenario Augmentation via \methodname{}}
\begin{figure*}[tb]
    \centering
    \includegraphics[width=0.95\textwidth]{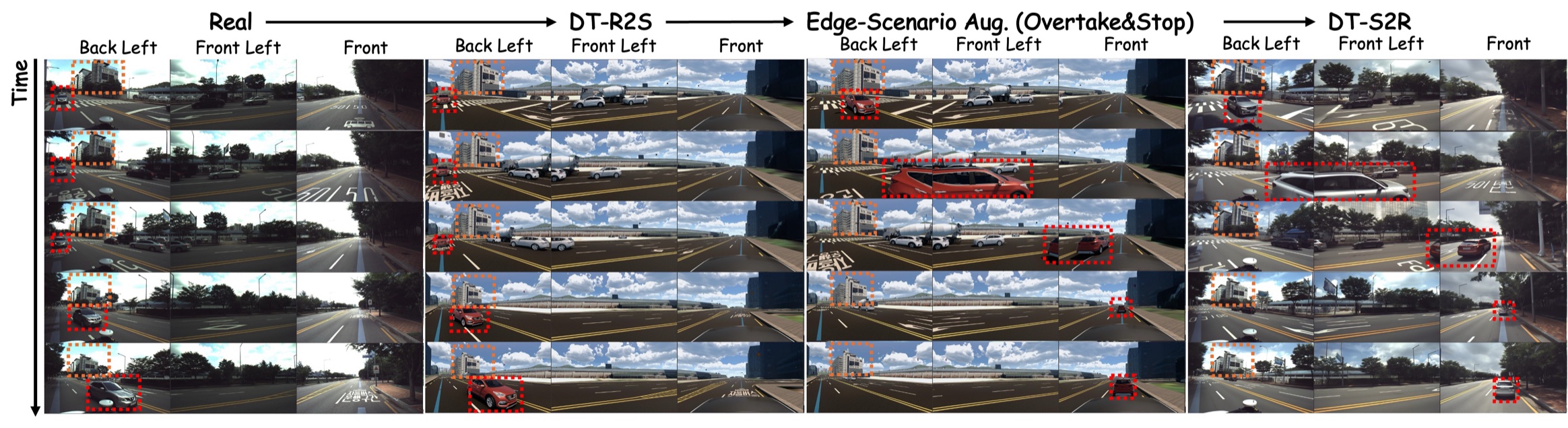}
    \caption{Preliminary attempt at long-tail mitigation utilizing \methodname{}. After reproducing a driving clip of \realpaired{}, a car showing normal behavior is edited to overtake and stop (red). The resulting images contain the augmented edge-case scenario, maintaining the original driving scenery (orange).}
    \label{fig:real2sim2real_scenario_edit}
\end{figure*}

Beyond normal scene reproduction, our framework can offer a mechanism to generate safety-critical data grounded in geo-referenced real-world environments. 
\reffigure{fig:real2sim2real_scenario_edit} demonstrates our exploratory step toward this direction.
By perturbing dynamic actors within a spatially anchored digital twin, we can synthesize hazardous edge cases directly from normal driving logs.

\begin{table*}[tb]
\centering
\caption{Evaluation of the efficacy that the generated synthetic data deliver during training as a proxy for a real-world target region \realpaired{}. Oracle is highlighted \blue{blue}, and the baseline measured using real-world off-target regions (\realunpaired{}) is colored \cyan{red}.}
\label{tab:region_without_real_3d}
\scriptsize
\renewcommand{\arraystretch}{1.45}
\setlength{\tabcolsep}{2.4pt}
\makebox[\columnwidth][c]{%
    \resizebox{0.8\textwidth}{!}{%
        \begin{tabular}{@{}*{4}{>{\centering\arraybackslash}m{0.085\textwidth}}*{5}{>{\centering\arraybackslash}m{0.048\textwidth}}*{5}{>{\centering\arraybackslash}m{0.048\textwidth}}@{}}
        \toprule
            \multicolumn{2}{c}{\textbf{Real}} & \multicolumn{2}{c}{\textbf{Synthetic}} & \multicolumn{5}{c}{\textbf{\detr{}}} & \multicolumn{5}{c}{\textbf{\petr{}}} \\
            \cmidrule(lr){1-2} \cmidrule(lr){3-4} \cmidrule(lr){5-9} \cmidrule(lr){10-14}
            \boldsymbol{\realpaired{}} & \boldsymbol{\realunpaired{}} & \boldsymbol{\simrealpaired{}} & \boldsymbol{\simrealadc{}} & \textbf{\map{}}$\uparrow$ & \textbf{\nds{}}$\uparrow$ & \textbf{0--15m}$\uparrow$ & \textbf{15--30m}$\uparrow$ & \textbf{30--50m}$\uparrow$ & \textbf{\map{}}$\uparrow$ & \textbf{\nds{}}$\uparrow$ & \textbf{0--15m}$\uparrow$ & \textbf{15--30m}$\uparrow$ & \textbf{30--50m}$\uparrow$ \\
        \midrule
            \rowcolor{oracleblue} \pmb{\checkmark{}} & \xmark{} & \xmark{} & \xmark{} & \underline{48.83} & \underline{54.39} & \textbf{86.88} & \textbf{58.63} & \underline{29.86} & \underline{55.23} & \textbf{58.10} & 88.27 & \underline{63.55} & \textbf{37.94} \\
        \midrule
            \rowcolor{BaselineRed} \xmark{} & \pmb{\checkmark{}} & \xmark{} & \xmark{} & 39.98 & 48.99 & 84.33 & 48.92 & 23.56 & 50.59 & 55.35 & \underline{89.55} & 58.42 & 32.31 \\
            \xmark{} & \xmark{} & \xmark{} & \pmb{\checkmark{}} & 29.39 & 39.86 & 61.05 & 36.52 & 15.25 & 34.41 & 41.29 & 62.47 & 41.46 & 17.96 \\
            \xmark{} & \xmark{} & \pmb{\checkmark{}} & \pmb{\checkmark{}} & 45.50 & 52.21 & 81.67 & 52.64 & 27.45 & 49.07 & 53.69 & 83.54 & 55.46 & 30.87 \\
            \rowcolor{tablegrey} \xmark{} & \pmb{\checkmark{}} & \pmb{\checkmark{}} & \pmb{\checkmark{}} & \textbf{51.02} & \textbf{55.77} & \underline{84.68} & \underline{57.89} & \textbf{33.55} & \textbf{55.79} & \underline{57.95} & \textbf{89.84} & \textbf{63.91} & \underline{37.57} \\
        \midrule
            \multicolumn{4}{c}{\textbf{Improvement over \blue{Oracle}}} & \improve{2.19} & \improve{1.38} & \degrade{2.20} & \degrade{0.74} & \improve{3.69} & \improve{0.56} & \degrade{0.15} & \improve{1.57} & \improve{0.36} & \degrade{0.37} \\
            \multicolumn{4}{c}{\textbf{Improvement over \cyan{Baseline}}} & \improve{11.04} & \improve{6.78} & \improve{0.35} & \improve{8.97} & \improve{9.99} & \improve{5.20} & \improve{2.60} & \improve{0.29} & \improve{5.49} & \improve{5.26} \\
        \bottomrule
        \end{tabular}
        }%
    }
\end{table*}

\section{Conclusion}
In this work, we investigated the value of georeferenced digital twins as a scalable, annotation-free training data source for camera-based 3D perception in a novel deployment scenario.  
By conditioning a diffusion model on rendered views, depth, and semantics from digital twin reconstructions, our \methodname{} framework synthesizes photorealistic driving data across both reconstructed scenes and novel simulator scenarios. 
Our experimental findings demonstrate that 3D detectors trained solely on this synthetic data recover a substantial portion of full performance of a target-region real-data oracle, and further outperform that oracle when co-trained with existing out-of-region real datasets. Where digital twin infrastructure is accessible, \methodname{} provides a practical and cost-effective pathway to scale 3D perception systems, which mitigates the expense of manual on-site data collection and annotation.

\begin{realblock}
\section*{Acknowledgments}
This research was supported by a grant(code 20014476) from R\&D Program funded by Ministry of Trade, Industry and Energy of Korean government. 
This research (paper) used datasets from \textit{High-precision data collection vehicle daytime city road data}. 
All data information can be accessed through \textit{AI-Hub} (\url{http://www.aihub.or.kr}).
\end{realblock}

\let\oldthebibliography\thebibliography
\renewcommand{\thebibliography}[1]{%
  \oldthebibliography{#1}%
  \setlength{\itemsep}{1pt}%
  \setlength{\parsep}{0pt}%
}
\bibliographystyle{IEEEtran}
\bibliography{IEEEabrv,bibfile}

\end{document}